\pdfoutput=1
\documentclass{article}

\usepackage[final]{colm2026_conference}

\usepackage{microtype}
\usepackage{graphicx}
\usepackage{booktabs}
\usepackage{amsmath}
\usepackage{amssymb}
\usepackage{xcolor}
\usepackage{multirow}
\usepackage{lineno}

\usepackage[hypertexnames=false]{hyperref}
\definecolor{darkblue}{rgb}{0, 0, 0.5}
\hypersetup{
  colorlinks=true,
  citecolor=darkblue,
  linkcolor=darkblue,
  urlcolor=darkblue,
  pdftitle={When Agents Commit Too Soon: Diagnosing Premature Commitment in LLM Agents},
  pdfauthor={Aman Mehta, Anupam Datta},
  pdfkeywords={LLM agents, representational commitment, hidden states, behavioral variability}
}

\usepackage{mathtools}
\usepackage{amsthm}
\usepackage{float}

\title{When Agents Commit Too Soon: Diagnosing\\Premature Commitment in LLM Agents}

\author{Aman Mehta \\
Snowflake AI Research \\
\texttt{aman.mehta@snowflake.com}
\And
Anupam Datta \\
Snowflake AI Research \\
\texttt{anupam.datta@snowflake.com}}

\begin{document}

\ifcolmsubmission
\linenumbers
\fi

\maketitle
\lhead{Workshop on Agent Behavior at COLM 2026}

\begin{abstract}
Tool-using LLM agents make early choices about what to search for, which evidence to trust, and what to do next. A wrong early choice can steer every later tool call, so an agent may produce a coherent but incorrect result. Existing evaluation does not reveal this process: final-answer scoring comes only after completion, and agreement across runs can be high even when every run is wrong. We ask whether hidden states reveal that repeated runs have already settled into stable behavior. Our contributions are threefold. First, we introduce \textbf{representational commitment}, which measures hidden-state similarity across repeated runs at the same step. On Llama-3.1-70B running ReAct on HotpotQA, we show that higher step-4 similarity is associated with lower whole-trajectory variability ($r=-0.35$); we find the same pattern in action sequences and replicate it on two other models and StrategyQA. This provides an internal measure for studying when agent behavior stabilizes. Second, we establish the measure's scope. Shared evidence explains much of the observed association, the HotpotQA result disappears on hard questions, and correct and confidently wrong questions overlap. We therefore show that the measure detects stability, not correctness, and cannot by itself establish that commitment was premature. Third, we evaluate monitoring and intervention. A monitor built from three partial runs predicts whole-trajectory variability (mean AUROC 0.81). A commitment prompt reduces variability by 28\% relative to a token-matched filler, but not significantly relative to the standard agent after correction, and it does not improve accuracy. Together, these contributions provide a measurement and evaluation framework for studying how agent trajectories stabilize and for testing when additional evidence, sampling, or review may be useful.
\end{abstract}

\section{Introduction}

LLM agents can chain reasoning steps, tool calls, and retrieved evidence over long tasks \citep{react, cot}. Such tasks increasingly include multi-step research, software edits, and web transactions. In this setting, reliability depends not only on the final answer but on whether the trajectory remains open to evidence. Agents sometimes settle on one interpretation early and follow it for the rest of the run. We call this \textbf{representational commitment}, and reserve \textbf{premature commitment} for the hazardous case in which early settling forecloses useful revision. Nothing crashes, and the trajectory can look coherent; the risk is that it may be coherent around the first reading, right or wrong.

This matters because final-answer checks and agreement checks observe outcomes rather than the process of settling. A single answer cannot reveal that the trajectory stabilized early. Cross-run agreement helps, but agreement is ambiguous: a confidently-wrong agent can agree with itself as reliably as a correct one. Prior work shows that behavioral divergence concentrates at early decision points \citep{mehta2026behavioral} and that repeated agreement can preserve silent semantic failure \citep{mehta2026confident}. The practical question is therefore not just whether agents agree, but whether the model has internally settled before the run is over.

We ask whether agents that behave similarly across repeated runs also converge internally before they finish. This design does not observe one run changing its mind. It measures whether repeated runs have converged by a given step.

\textbf{Practical motivation.} A settling signal could direct more sampling, evidence gathering, or human review toward inputs whose partial runs remain unstable. Conversely, converged runs might avoid redundant sampling, but only after a separate verifier checks their validity. We therefore separate two questions that output confidence often conflates: \emph{has the process stabilized?} and \emph{is the result correct?}

\textbf{Operational definition.} We define \emph{representational commitment} as cross-run hidden-state convergence at a fixed agent step: run the same input $n$ times at non-zero temperature, take the hidden state at the last token of step $s$, and measure mean pairwise cosine similarity across runs. High similarity indicates that the runs have reached similar representations; it may arise from shared observations, shared processing, or both. \emph{Activation similarity} is the measurement; \emph{representational commitment} is the construct it is intended to index.

\textbf{Contributions.}
\begin{enumerate}
    \item \textbf{A measure of behavioral settling.} We introduce representational commitment, the similarity of hidden states across repeated runs at the same step. We show that higher similarity is associated with lower variation in total steps and action sequences on Llama-HotpotQA, and we replicate the association across models and on another benchmark.
    \item \textbf{A clear account of what the measure captures.} We show that shared retrieved evidence explains much of the association, that the result is weak on hard HotpotQA questions, and that correct and confidently wrong behavior can look equally stable. The measure detects stability, not correctness or whether settling was premature.
    \item \textbf{An evaluation of monitoring and intervention.} We show that repeated partial runs can predict whole-trajectory variability, and we compare a commitment prompt with a token-matched filler. These tests identify a possible role in allocating additional sampling or review, while the lack of an accuracy gain demonstrates the need for a separate verifier.
\end{enumerate}

The paper follows the same sequence. We first establish the association and locate it in time and depth. We then test simple explanations and correctness boundaries. Finally, we examine replication, early detection, and a matched prompt comparison.

\section{Related work}

\textbf{Behavioral consistency in agents.} \citet{mehta2026behavioral} introduced multi-run behavioral consistency as an agent uncertainty signal on HotpotQA, while \citet{mehta2026confident} showed that repeated agreement can coexist with silent semantic failure on SWE-bench. Self-consistency \citep{wang2023selfconsistency} leverages output-level variance via majority voting but does not examine internal representations. We ask whether behavioral stability has a measurable signature in hidden states without treating stability itself as evidence of correctness.

\textbf{Probing LLM representations.} Hidden states can separate true from false statements in controlled settings \citep{burns2023discovering, marks2024geometry}, and language models can report confidence about their answers \citep{kadavath2022language}. Probes can also detect incorrect generations from internal states in single-turn settings \citep{azaria2023internal}, including in reasoning models \citep{zhang2025reasoning}. Standard probing asks \emph{what} one hidden state encodes \citep{belinkov2022probing}. We instead ask whether representations agree across independent trajectories. This requires comparing runs rather than scoring one state at a time.

\textbf{Representation engineering.} \citet{zou2023repe} introduced representation reading and control, and \citet{turner2023steering} demonstrated activation steering for behavioral modification. \citet{anthropic_persona} extracted ``persona vectors'' encoding character traits, and \citet{anthropic_axis} identified an ``assistant axis'' governing default model behavior. We ask whether commitment has a similarly measurable representational signature, without assuming that it forms a single linear axis.

\section{Method}

\textbf{Analysis overview.} For each question, we compare several independent runs at an intermediate step and again after completion. The first comparison measures hidden-state similarity; the second measures whole-trajectory behavioral stability. Our central test asks whether the former is associated with the latter.

\subsection{Setup and terminology}
\label{sec:setup}

\textbf{Agent step.} We use a ReAct \citep{react} agent that iterates Thought $\rightarrow$ Action $\rightarrow$ Observation. One such triple is one \emph{step}; ``step~4'' is the fourth triple in a run. A run ends when the agent calls \texttt{Finish} or hits the 25-step cap.

\textbf{What we extract.} At step $s$ we take the hidden state at the \emph{last output-token position} of that step, computed over the cumulative context (system instructions, the question, the $s{-}1$ prior triples, and the current Thought and Action). The last-token choice follows the single-turn probing convention \citep{burns2023discovering, marks2024geometry}; we treat it as a convention rather than a proven optimum and flag the comparison against pooled or trajectory-level representations as open.

\textbf{Activation similarity vs.\ representational commitment.} \emph{Activation similarity} is the measured quantity (Equation~\ref{eq:sim}, mean cross-run cosine of the last-token hidden state). \emph{Representational commitment} is the construct it is meant to index. We keep the two terms distinct throughout.

\textbf{Behavioral variability.} We use two complementary, deliberately coarse metrics across the 10 runs of a question. The primary metric is the coefficient of variation (CV, standard deviation divided by mean) of step counts. It is scale-invariant, so runs of 12 and 13 steps count as less variable than runs of 5 and 25 regardless of the mean. The second is action-sequence diversity, the proportion of unique action sequences among runs, which captures path identity. Lower values on both indicate more stable behavior, and they agree at $r = 0.71$. Neither establishes semantic equivalence: distinct trajectories can have the same length, and superficially different actions can implement similar reasoning. We therefore report them as behavioral proxies rather than a complete measure of trajectory consistency.

\textbf{Agent and task.} We deploy Llama-3.1-70B-Instruct \citep{llama3} on HotpotQA \citep{hotpotqa} with Search, Retrieve, and Finish tools. We select 100 validation questions: 50 ``easy'' (comparison questions with yes/no answers) and 50 ``hard'' (multi-hop, $\geq$2 retrieval steps). Each question is run 10 times at temperature $T{=}0.5$\footnote{$T{=}0.5$ yields useful behavioral variation. At $T{=}0$, repeated runs are identical; at $T{=}1.0$, they are much noisier. We did not run a full temperature sweep.} with a 25-step cap, giving 988 trajectories.\footnote{12 runs were excluded because hidden-state extraction was incomplete.} At step~4, 94 of 100 questions have at least two valid hidden states and enter the primary analysis. The other six are hard questions whose runs ended before enough step-4 states were recorded.

\textbf{Hidden-state extraction.} The model runs on 8$\times$80GB GPUs with pipeline parallelism (10 layers per GPU). At every step we record the hidden state at the last-token position from layers $\ell \in \{0, 8, 16, \ldots, 80\}$, giving $\mathbf{h}_{\ell}^{(s)} \in \mathbb{R}^{8192}$ per step $s$ and layer $\ell$.

\textbf{Activation similarity.} For question $q$ at step $s$, layer $\ell$, we compute the mean pairwise cosine similarity across runs:
\begin{equation}
\label{eq:sim}
\text{Sim}_q^{(s,\ell)} = \binom{n}{2}^{-1} \sum_{i < j} \cos\!\left(\mathbf{h}_{\ell,i}^{(s)},\; \mathbf{h}_{\ell,j}^{(s)}\right),
\end{equation}
where $n$ is the number of runs with hidden states at step $s$. We report the Pearson correlation $r$ between $\text{Sim}_q^{(s,\ell)}$ and $\text{CV}_q$ across questions. With 94 eligible questions at step~4, the analysis has about 80\% power to detect $|r| \gtrsim 0.29$ (two-tailed $\alpha=0.05$).

\textbf{Statistical methods and analysis status.} We use Pearson and partial Pearson correlations (difficulty and accuracy as covariates), paired $t$-tests for the intervention, bootstrap mediation \citep{preacher2008asymptotic}, LOOCV AUROC for the monitor, and 10{,}000-iteration permutation tests with Bonferroni correction over the step$\times$layer grid. Pearson correlation intervals use Fisher's $z$ transform; the partial-correlation interval adjusts for its two covariates. Bootstrap intervals are labeled explicitly. The step-4/layer-40 result was discovered in a 55-cell scan (5 steps $\times$ 11 layers) on Llama-HotpotQA, so it is exploratory. Before the replication runs, we fixed the expected direction of association. Qwen and Phi-3 use new trajectories on the same HotpotQA questions; StrategyQA uses a new benchmark. Model- and task-specific peak locations remain exploratory.

\section{Results}

\begin{figure*}[t]
\centering
\includegraphics[width=0.94\textwidth]{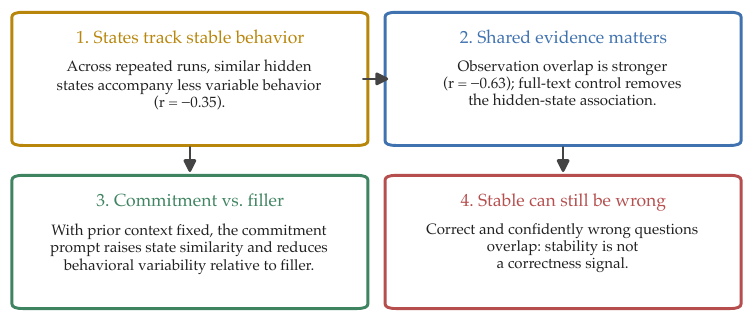}
\caption{Evidence map. Similar hidden states accompany more stable behavior, but shared evidence explains much of this association. With prior context fixed, the commitment and filler prompts produce different state similarity and behavioral variability. Neither result establishes correctness.}
\label{fig:evidence_map}
\end{figure*}

Figure~\ref{fig:evidence_map} summarizes the argument. We first ask whether convergence tracks stable behavior and where the association appears (\S\ref{sec:res-sim}--\ref{sec:temporal}). We then test whether shared evidence explains it and whether it says anything about correctness (\S\ref{sec:baselines}--\ref{sec:boundaries}). Finally, we examine replication, early detection, and a matched prompt comparison (\S\ref{sec:crossmodel}--\ref{sec:intervention}).

\subsection{A hidden-state signature of behavioral stability}
\label{sec:res-sim}

\begin{figure}[t]
\centering
\includegraphics[width=0.78\columnwidth]{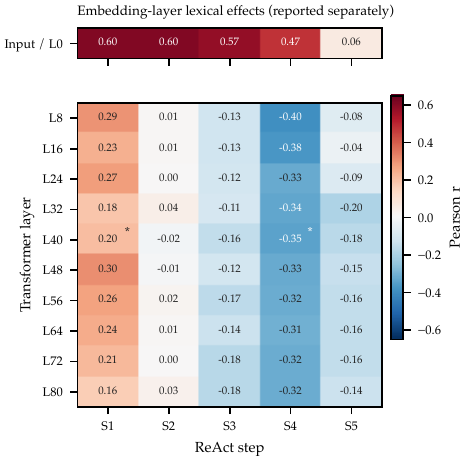}
\caption{Pearson $r$ between activation similarity and behavioral CV. Eligible questions vary by step (layer~40: $n=99,99,99,94,65$ for steps 1--5). Layer~0 is shown separately because lexical effects may dominate there. Asterisks mark cells with an available two-sided Pearson $p<.05$; an unmarked cell is not necessarily nonsignificant. The negative pattern concentrates at step~4.}
\label{fig:heatmap}
\end{figure}

Activation similarity at step~4 is negatively correlated with step-count CV: questions whose hidden states converge across runs have less variable trajectory lengths. At our primary mid-layer readout, layer~40, $r=-0.348$ (95\% CI $[-0.51,-0.16]$, $p=0.0006$; Figure~\ref{fig:heatmap}). The association spans all non-embedding layers. Its largest unadjusted magnitude occurs earlier, at layer~8 ($r=-0.40$); within layers 32--80, layer~40 has the largest magnitude. The top layer-40 similarity quartile has $4.1\times$ lower mean CV than the bottom quartile ($d=1.01$), and action-sequence diversity has the same ordering. This is an association with behavioral stability, not semantic identity.

The result survives multiple-comparison controls. A 10{,}000-iteration permutation test is significant at all seven layers from 32 to 80 ($p<0.002$; Appendix~\ref{app:permutation}). The step-4/layer-40 cell also survives Bonferroni correction over all 55 step$\times$layer cells (corrected $\alpha=0.0009$; permutation $p=0.0003$). The effect then reappears in two other models at different strongest layers (Section~\ref{sec:crossmodel}).

\subsection{When and where the association appears}
\label{sec:temporal}

We next ask whether the association appears throughout a run or at a particular point, and whether task difficulty explains it.

\textbf{Temporal profile.} The commitment-associated \emph{negative} signal is absent at steps~1--2, strengthens at step~3, peaks at step~4 ($r = -0.348$), and weakens at step~5 (Figure~\ref{fig:step_progression}, Appendix~\ref{app:temporal}). Some step-1 cells are positive (layer~40: $r=+0.20$, unadjusted $p=.048$); these are directionally opposite to the hypothesized convergence--consistency relationship.

\textbf{Task difficulty.} Both activation similarity and CV could be driven by task difficulty. A partial correlation controlling for accuracy and difficulty label strengthens the signal from $r = -0.35$ to $r = -0.45$ (95\% CI $[-0.60, -0.26]$, $p < 10^{-5}$) at layer~40, with all layers 32--80 strengthening (Appendix~\ref{app:partial}). This control does not address retrieval overlap, which is the more consequential confound below.

\subsection{Shared evidence explains much of the association}
\label{sec:baselines}

We test a key alternative explanation: runs that retrieve the same evidence may have both similar hidden states and similar behavior. We therefore compare hidden-state similarity against evidence-overlap measures and simpler predictors of step-count CV (Figure~\ref{fig:predictor_comparison}; full table in Appendix~\ref{app:baselines}). Question length ($r = 0.10$, $p = .31$), context size ($r = 0.04$, $p = .66$), and step-3 thought length ($r = 0.12$, $p = .24$) do not predict CV.

\textbf{Observation overlap is stronger.} Runs that read similar evidence will have both similar hidden states and similar behavior. Observation TF-IDF similarity predicts CV at $r=-0.63$ and document Jaccard at $r=-0.40$, both stronger than hidden-state similarity ($r=-0.35$); search-query overlap reaches $r=-0.24$ ($n=94$). Controlling for document identity leaves a partial association ($r=-0.31$, $p=.003$), and within the top TF-IDF quartile it remains ($r=-0.47$, $p=.011$, $n=28$). However, conditioning on full-text TF-IDF absorbs it (partial $r=0.09$, n.s.). Thus, much of the association in unmodified runs has a simple explanation: \emph{similar inputs produce similar states and behavior}. The intervention in \S\ref{sec:intervention} provides a separate test by holding the prior context fixed, but it does not identify which wording change causes the effect.

\begin{figure}[t]
\centering
\includegraphics[width=\columnwidth]{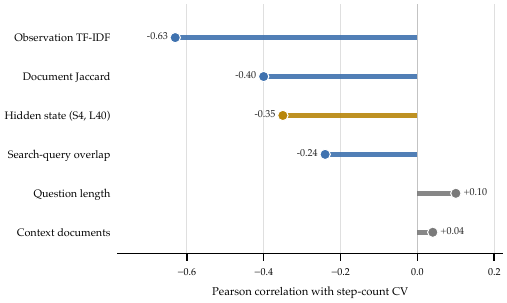}
\caption{Naturalistic predictors of step-count CV. Observation-overlap measures outperform hidden-state similarity. The highlighted activation result remains predictive before full-text control, but full-text TF-IDF absorbs it.}
\label{fig:predictor_comparison}
\end{figure}

\subsection{Settling does not imply correctness}
\label{sec:boundaries}

We next test whether behavioral stability also predicts correctness. It does not.

\textbf{Per-run correctness.} Linear probes fail to predict per-run correctness from hidden states (AUC 0.34--0.56 across all configurations, including single-turn generation), because correctness here is largely question-determined (64 questions are 10/10 correct, 23 are 0/10). The consistency signal is a \emph{between-run relational} property, not a within-run absolute one.

\textbf{Hard vs.\ easy questions.} At layer~40, the signal is carried by easy questions ($r=-0.57$, $p=1.65\times10^{-5}$) rather than hard questions ($r=-0.02$, $p=0.88$; Figure~\ref{fig:hard_vs_easy}, Appendix~\ref{app:hardeasy}). This sharply limits the practical scope of the HotpotQA result. One possibility is that low-accuracy hard questions can still have high similarity, but the data do not establish why the association disappears.

\begin{figure}[t]
\centering
\includegraphics[width=\columnwidth]{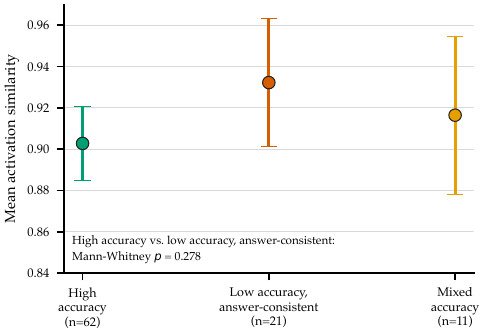}
\caption{Llama layer-40 activation similarity at step~4 by descriptive group. Points show group means; error bars are 95\% normal-approximation confidence intervals of the mean ($1.96\,\mathrm{SD}/\sqrt{n}$). The high-accuracy and confidently-wrong (low-accuracy, answer-consistent) groups overlap (Mann--Whitney $p=.278$). The asymmetric, post hoc groups are descriptive rather than evidence of equivalence.}
\label{fig:commitment_categories}
\end{figure}

Across all 94 eligible questions, we form three descriptive groups: high accuracy (accuracy $\geq0.8$, $n=62$), low accuracy but answer-consistent (accuracy $\leq0.2$ with a majority-answer frequency $\geq0.6$, $n=21$), and mixed accuracy ($n=11$). The second group is the confidently-wrong case; it is \emph{not} a low-variability group in the step-count sense, since its mean CV ($0.235$) exceeds the high-accuracy group's ($0.112$). At Llama layer~40, the first two groups have overlapping mean activation similarity ($0.903$ vs.\ $0.932$, Mann--Whitney $p=0.278$). We use this only as descriptive evidence that similarity does not directly encode correctness (Figure~\ref{fig:commitment_categories}). We do not report the same comparison for Qwen because the required per-question distributions were not retained.

\textbf{Answer agreement.} Activation similarity does not predict cross-run answer agreement ($r = -0.13$, $p = 0.22$). The signal tracks how agents \emph{reach} answers, not whether the answers match, consistent with its trajectory-level target (\S\ref{sec:setup}).

\subsection{The association recurs across models and tasks}
\label{sec:crossmodel}

We evaluate whether the association generalizes beyond the original model and dataset. We replicate on Qwen-2.5-72B-Instruct \citep{qwen2.5} and Phi-3-Medium-14B \citep{phi3}. Each model runs the same 100 HotpotQA questions 10 times (988 Llama, 998 Qwen, and 1{,}000 Phi-3 trajectories).

\begin{figure}[t]
\centering
\includegraphics[width=\columnwidth]{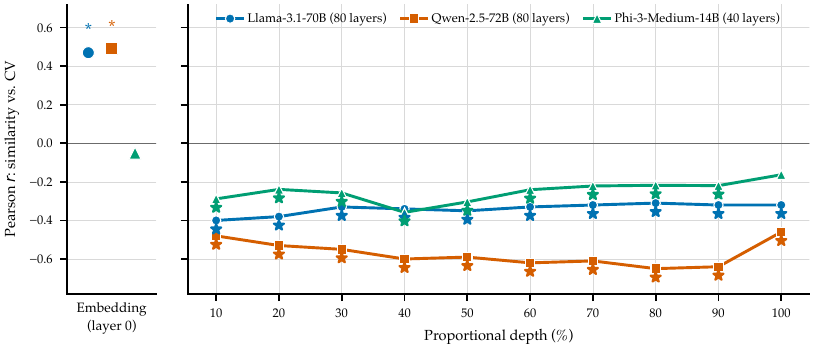}
\caption{Layer-wise correlation (activation similarity vs.\ CV) at step~4 across three models, layers normalized to proportional depth. The embedding layer is plotted in the separate narrow panel at left because its positive association is a lexical artifact (Appendix~\ref{app:crossmodel}) that would otherwise stretch the shared axis. Marker shape and color both identify the model; an asterisk beside or below a marker denotes unadjusted $p<.05$.}
\label{fig:crossmodel}
\end{figure}

The negative correlation replicates in both models. Qwen's largest step-4 magnitude is $r=-0.65$ at layer~64 (95\% CI $[-0.75,-0.51]$, $p<10^{-6}$, $n=90$). Phi-3 reaches $r=-0.36$ at step~4, layer~16 (95\% CI $[-0.53,-0.16]$, $p=0.0005$, $n=91$), and its strongest reported result at the fixed layer-16 readout occurs at step~5 ($r=-0.58$, $p<10^{-6}$, $n=87$). The strongest locations differ across models, so the depth profile is not fixed (Figure~\ref{fig:crossmodel}; full results in Table~\ref{tab:crossmodel_full}, Appendix~\ref{app:crossmodel}).

\textbf{Cross-benchmark generalization on Llama: StrategyQA.}
\label{sec:strategyqa}

Holding the model fixed at Llama, we replicate on StrategyQA \citep{strategyqa} (implicit multi-step reasoning, yes/no answers; 50 questions balanced by answer, $\geq$2 decomposition steps, 10 runs each, $T{=}0.5$). The reported peak is at step~3: $r=-0.83$ (95\% CI $[-0.90,-0.72]$, $p<10^{-13}$). The association spans layers 8--80 ($|r|>0.72$, $p<10^{-8}$) and plateaus across layers 56--72. This peak occurs one step earlier than on HotpotQA, whose reasoning chains are longer on average (about 12 steps vs.\ 5.1). Controlling for accuracy leaves the correlation unchanged ($r=-0.83$). Together, the HotpotQA and StrategyQA results test two replication axes: architecture and benchmark. We did not run Qwen or Phi-3 on StrategyQA.

\subsection{Using partial runs to predict behavioral variability}
\label{sec:runtime}

\textbf{Runtime detection.} Can a monitor use repeated partial runs to identify behaviorally stable or variable questions before the runs finish? We train a logistic classifier from step-4 hidden states (LOOCV; five hidden-state features against surface baselines; full results in Appendix~\ref{app:runtime}).

Hidden-state features discriminate well. A per-layer similarity profile reaches AUROC~0.97 on Llama under quintile labeling, which omits the ambiguous middle 60\% and is the easier task. Under a median split, the best feature reaches 0.85 on Llama and 0.88 on Qwen. Question length is available before any run and stays near chance (0.52--0.65). Context-document count is also available early, but its score direction was chosen after observing the labels; we therefore do not treat it as a fair comparison. Other surface statistics are available only after completion (Appendix~\ref{app:runtime}). With three runs, the monitor reaches mean AUROC $0.81$ (subsampling SD $0.07$; 95\% bootstrap CI $[0.68,0.94]$).

\subsection{Comparing a commitment prompt with a matched filler}
\label{sec:intervention}

The observational analysis cannot disentangle convergent evidence from convergent internal processing. We therefore test a prompting intervention at step~3, appending an instruction to commit to a reasoning strategy before continuing (Appendix~\ref{app:prompts}). All conditions share the same prior context. The control is the standard ReAct agent, and a semantically neutral filler is matched for token length. This is still a coarse intervention: commitment changes lexical content, imperative voice, and strategy framing together, while filler controls only token count.

We run all three conditions on 100 HotpotQA questions with 10 runs each on Llama-3.1-70B ($T = 0.5$); Table~\ref{tab:intervention} reports the results. Accuracy uses corrected matching (substring or token-level F1 $\geq 0.5$ against gold).

\begin{table}[ht]
\centering
\small
\begin{tabular}{lccc}
\toprule
\textbf{Metric} & \textbf{Control} & \textbf{Filler} & \textbf{Commitment} \\
\midrule
Behavioral CV & 0.112 & 0.132 & 0.095 \\
Action seq.\ diversity & 0.250 & 0.295 & 0.223 \\
Accuracy\textsuperscript{\dag} & 0.923 & 0.902 & 0.908 \\
\midrule
\multicolumn{4}{l}{\emph{Pairwise comparisons (paired $t$-test; raw $p$; Holm $p$ shown for CV):}} \\
\midrule
 & \multicolumn{3}{c}{\textbf{Filler vs.\ Commitment} (token-controlled)} \\
CV reduction & \multicolumn{3}{c}{28\% ($d = 0.33$, $p = .001$, Holm $p = .003$)} \\
Diversity reduction & \multicolumn{3}{c}{24\% ($d = 0.47$, $p < .001$)} \\
Accuracy difference & \multicolumn{3}{c}{n.s.\ ($p = .32$)} \\
\midrule
 & \multicolumn{3}{c}{\textbf{Control vs.\ Commitment} (net over standard agent)} \\
CV reduction & \multicolumn{3}{c}{15\% ($d = 0.20$, $p = .044$, Holm $p = .088$, n.s.)} \\
\midrule
 & \multicolumn{3}{c}{\textbf{Control vs.\ Filler}} \\
CV change & \multicolumn{3}{c}{+18\% ($d = 0.18$, $p = .071$, n.s.)} \\
\bottomrule
\end{tabular}
\caption{Three-condition prompting intervention ($n=100$ questions, 10 runs each). The token-count-controlled comparison (Filler vs.\ Commitment) survives Holm correction over the three CV comparisons; the net effect relative to Control does not. \textsuperscript{\dag}Accuracy uses corrected matching (substring or token-level F1 $\geq0.5$ against gold).}
\label{tab:intervention}
\end{table}

The filler does not reduce CV relative to control ($p=.071$), so extra tokens alone do not help. Commitment cuts CV by 28\% relative to filler (paired $t(99)=3.29$, $p=.001$, $d=0.33$, Holm $p=.003$), and action-sequence diversity falls by 24\% ($d=0.47$, $p<.001$). Compared with the standard agent, the CV reduction is smaller (15\%, Holm $p=.088$) and does not survive correction. The reported filler-versus-commitment accuracy difference is not significant ($p=.32$). Thus commitment framing reduces measured behavioral variability relative to a token-count-matched insertion, but we do not find evidence that it improves accuracy.

\textbf{Representational-level evidence.} For 89 questions with valid step-4 states in all three conditions, the prior trajectory is shared through step~3; 11 questions lack a complete matched comparison. At step~4, commitment is highest (0.995), filler intermediate (0.979), and control lowest (0.922) (Appendix~\ref{app:intervention_decomposition}, Figure~\ref{fig:intervention_step_progression}). Commitment increases convergence beyond the token-count-matched filler ($\Delta=+0.016$, $d=0.97$, $p<.001$, layer~40). The filler raises similarity (0.922 to 0.979) but worsens CV (0.112 to 0.132). This shows that raw similarity can rise from shared text without producing commitment's behavioral pattern.

\begin{figure}[t]
\centering
\includegraphics[width=\columnwidth]{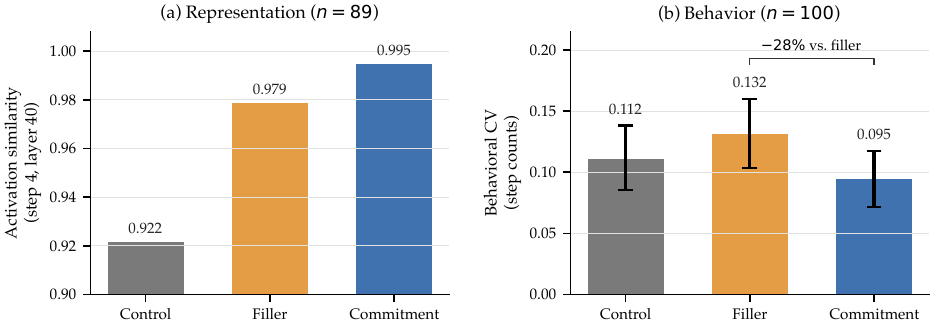}
\caption{The intervention separates representational similarity from token count. \textbf{(A)} Among the 89 questions matched across all conditions, activation similarity rises from Control to Filler to Commitment. The y-axis is zoomed. Question-level activation uncertainty is unavailable, so Panel~A shows point estimates only. \textbf{(B)} Behavioral CV uses all 100 questions; error bars are 95\% normal-approximation confidence intervals for the mean.}
\label{fig:intervention_bars}
\end{figure}

\textbf{Mediation analysis.} A mediation analysis finds that activation similarity accounts for part of the change in behavioral variability, but its direct and indirect paths point in opposite directions (Appendix~\ref{app:intervention_decomposition}). We therefore treat it as a statistical decomposition, not evidence of a causal mechanism.

The effect concentrates on originally variable questions ($d=0.44$, $p=.018$). Among all 100 questions, 89 are consistent-correct under Control; 88 of those remain consistent-correct under Commitment (Appendix~\ref{app:intervention_decomposition}). This decomposition is descriptive and does not establish that commitment is risk-free. The single-layer steering attempt (Appendix~\ref{app:steering}) gave mixed results. Table~\ref{tab:summary} consolidates all results.

\section{Discussion}

\textbf{Interpretation.} We interpret representational commitment as settling, not failure. It becomes premature only when the agent settles before gathering or using evidence that would change its path. Our experiments do not directly observe such a rejected correction. The overlap between high- and low-accuracy groups shows why settling cannot be labeled a failure without an external correctness check.

\textbf{Practical implications.} Our results suggest that the signal could help allocate compute and review. Inputs whose partial runs remain unstable may benefit from more samples or more evidence. Converged runs may be candidates for early exit, but only after a separate verifier checks validity. We do not recommend the commitment prompt as a general deployment intervention because it can reinforce a wrong interpretation.

\textbf{Mechanistic scope.} Recent work finds linear directions for persona \citep{anthropic_persona}, default behavior \citep{anthropic_axis}, and truth \citep{marks2024geometry}. Commitment may also have representational structure, but our evidence is predictive rather than mechanistic. Its strongest location varies by task and architecture, and our exploratory direction does not isolate a commitment-specific axis. Unlike self-consistency \citep{wang2023selfconsistency}, our measurement reads hidden states before completion.

\textbf{Safety and ethics.} Stable behavior can still be wrong. A monitor that triggers early exit, or a prompt that encourages commitment, could make an agent persist in a harmful interpretation. Neither should be used to approve consequential actions without an independent verifier and human oversight where appropriate.

\textbf{Limitations.} In unmodified runs, observation TF-IDF predicts behavior better than activation similarity, and controlling for full-text overlap removes the activation association. Step-count CV and action diversity are coarse measures: different trajectories can still have the same length or action pattern. We use the last token at each step, but pooled-token or whole-trajectory representations may work differently. The HotpotQA association vanishes on hard questions, the strongest step and layer vary by task and model, and all runs use one temperature. We have not tested code, math, or embodied tasks. The prompt intervention changes several aspects of the text at once, the mediation result is mixed, and single-layer steering is unreliable. Our claim is predictive, not an assertion that hidden states faithfully explain the model's reasoning \citep{turpin2023language}.

\textbf{Future directions.} The most informative next steps are (1) a replayed-observation experiment that holds retrieved documents fixed while letting reasoning vary, to separate processing from retrieval; (2) equivalence testing (TOST) for the correctness-agnostic claim; and (3) a fully specified hidden-state compute router evaluated against output-based adaptive consistency \citep{aggarwal2023adaptive}. Multi-layer steering and circuit tracing at divergence points \citep{zou2023repe, anthropic_circuits} are natural mechanistic follow-ups.

\section{Conclusion}

Reliable agents require two separate judgments: has the reasoning process stabilized, and is its conclusion valid? We show that cross-run hidden-state convergence carries information about the first. The association recurs across models and tasks, and the commitment and filler prompts produce different similarity and variability after the same prior context. However, the association in unmodified runs partly reflects shared retrieval, disappears on hard HotpotQA questions, and does not distinguish right from wrong. Representational commitment may therefore help target resampling, evidence gathering, or review; it cannot replace an independent correctness check.


\bibliography{references}
\bibliographystyle{colm2026_conference}

\appendix

\section{Preliminary steering experiment}
\label{app:steering}

We ran a preliminary steering test on five questions. At step~4, layer~40, we formed a direction from the difference between states drawn from low-CV, high-accuracy questions and states drawn from higher-CV questions. We then added this direction during inference with scale $\alpha=1.5$. Steering reduced step-count CV on two questions (0.31 to 0.18 and 0.27 to 0.12). On the other three, it had no effect or slightly increased variance, and it sometimes disrupted reasoning. This small experiment does not establish a reliable control method. One possible explanation is timing: the prompt changes generation from step~3 onward, while single-layer steering acts later and in only one subspace. Multi-layer or multi-step steering \citep{zou2023repe} would require a separate evaluation.

\section{Temporal profile}
\label{app:temporal}

\begin{figure}[H]
\centering
\includegraphics[width=0.78\columnwidth]{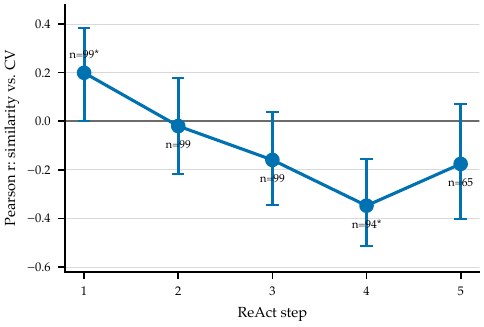}
\caption{Layer-40 Pearson correlation between activation similarity and CV across steps, using only questions with at least two valid hidden states at the plotted step. Labels report eligible $n$ ($99,99,99,94,65$); intervals are 95\% Fisher-$z$ confidence intervals. Asterisks denote unadjusted two-sided $p<.05$. The negative commitment-associated signal peaks at step~4.}
\label{fig:step_progression}
\end{figure}

\section{Permutation test results}
\label{app:permutation}

\begin{table}[H]
\centering
\small
\begin{tabular}{lccc}
\toprule
Layer & $r$ & Param.\ $p$ & Perm.\ $p$ \\
\midrule
32 & $-0.340$ & $0.0008$ & $0.0007$ \\
\textbf{40} & $\mathbf{-0.348}$ & $\mathbf{0.0006}$ & $\mathbf{0.0003}$ \\
48 & $-0.326$ & $0.0013$ & $0.0012$ \\
56 & $-0.320$ & $0.0017$ & $0.0011$ \\
64 & $-0.313$ & $0.0021$ & $0.0014$ \\
72 & $-0.316$ & $0.0019$ & $0.0019$ \\
80 & $-0.319$ & $0.0017$ & $0.0019$ \\
\bottomrule
\end{tabular}
\caption{Permutation test at step~4 (10{,}000 iterations). All layers 32--80 are significant at $p < 0.002$.}
\label{tab:permutation}
\end{table}

\section{Partial correlation results}
\label{app:partial}

\begin{table}[H]
\centering
\small
\begin{tabular}{lcccc}
\toprule
Layer & Raw $r$ & Raw $p$ & Partial $r$ & Partial $p$ \\
\midrule
32 & $-0.340$ & $.0008$ & $-0.456$ & $< .0001$ \\
\textbf{40} & $\mathbf{-0.348}$ & $\mathbf{.0006}$ & $\mathbf{-0.445}$ & $\mathbf{< .0001}$ \\
48 & $-0.326$ & $.0013$ & $-0.416$ & $< .0001$ \\
56 & $-0.320$ & $.0017$ & $-0.399$ & $< .0001$ \\
64 & $-0.313$ & $.0021$ & $-0.387$ & $.0001$ \\
72 & $-0.316$ & $.0019$ & $-0.387$ & $.0001$ \\
80 & $-0.319$ & $.0017$ & $-0.375$ & $.0002$ \\
\bottomrule
\end{tabular}
\caption{Partial correlations at step~4, controlling for accuracy and difficulty label. The signal strengthens after removing difficulty-related variance.}
\label{tab:partial}
\end{table}

\section{Full cross-model layer-wise results}
\label{app:crossmodel}

Table~\ref{tab:crossmodel_full} reports the full layer-wise correlations at step~4 for all three models. Phi-3 layers are mapped to proportionally equivalent depths (e.g., Phi-3 layer~16 $\approx$ 40\% depth).

\textbf{Layer-0 lexical effects.} Layer~0 is the embedding-layer state at the last generated token. Its positive association for Llama ($r=+0.47$) and Qwen ($r=+0.49$) may reflect repeated terminal tokens or other lexical effects. We therefore report it separately and do not interpret it as commitment. Phi-3 has no corresponding association ($r=-0.05$, $p=.63$).

\begin{table}[H]
\centering
\scriptsize
\setlength{\tabcolsep}{3pt}
\begin{tabular}{lcccccc}
\toprule
 & \multicolumn{2}{c}{Llama 3.1 70B} & \multicolumn{2}{c}{Qwen 2.5 72B} & \multicolumn{2}{c}{Phi-3 14B} \\
\cmidrule(lr){2-3} \cmidrule(lr){4-5} \cmidrule(lr){6-7}
Layer & $r$ & $p$ & $r$ & $p$ & $r$ & $p$ \\
\midrule
0  & $0.47^*$ & $< .001$ & $0.49^*$ & $< .001$ & $-0.05$ & $.634$ \\
8  & $\mathbf{-0.40^*}$ & $\mathbf{< .001}$ & $-0.48^*$ & $< .001$ & $-0.24^*$ & $.023$ \\
16 & $-0.38^*$ & $< .001$ & $-0.53^*$ & $< .001$ & $\mathbf{-0.36^*}$ & $\mathbf{< .001}$ \\
24 & $-0.33^*$ & $.001$ & $-0.55^*$ & $< .001$ & $-0.24^*$ & $.022$ \\
32 & $-0.34^*$ & $< .001$ & $-0.60^*$ & $< .001$ & $-0.22^*$ & $.038$ \\
$40^\dagger$ & $-0.35^*$ & $< .001$ & $-0.59^*$ & $< .001$ & $-0.16$ & $.123$ \\
48 & $-0.33^*$ & $.001$ & $-0.62^*$ & $< .001$ & -- & -- \\
56 & $-0.32^*$ & $.002$ & $-0.61^*$ & $< .001$ & -- & -- \\
\textbf{64} & $-0.31^*$ & $.002$ & $\mathbf{-0.65^*}$ & $\mathbf{< .001}$ & -- & -- \\
72 & $-0.32^*$ & $.002$ & $-0.64^*$ & $< .001$ & -- & -- \\
80 & $-0.32^*$ & $.002$ & $-0.46^*$ & $< .001$ & -- & -- \\
\midrule
Unadj.\ sig.\ non-embed.\ layers & \multicolumn{2}{c}{10 / 10} & \multicolumn{2}{c}{10 / 10} & \multicolumn{2}{c}{4 / 5} \\
Largest $|r|$ & \multicolumn{2}{c}{$-0.40$ (L8)} & \multicolumn{2}{c}{$-0.65$ (L64)} & \multicolumn{2}{c}{$-0.36$ (L16)} \\
Accuracy & \multicolumn{2}{c}{0.70} & \multicolumn{2}{c}{0.81} & \multicolumn{2}{c}{0.69} \\
\bottomrule
\end{tabular}
\caption{Cross-architecture validation at step~4 on HotpotQA. All three models show negative correlations outside the embedding layer. Phi-3's strongest reported result is one step later, at step~5 ($r=-0.58$; Section~\ref{sec:crossmodel}). $^*p<0.05$ is unadjusted. $^\dagger$Layer~40 is the primary Llama readout, while layer~8 has the largest unadjusted magnitude. Accuracy here uses substring-only matching; Table~\ref{tab:intervention} uses corrected matching.}
\label{tab:crossmodel_full}
\end{table}

\begin{figure}[H]
\centering
\includegraphics[width=0.8\columnwidth]{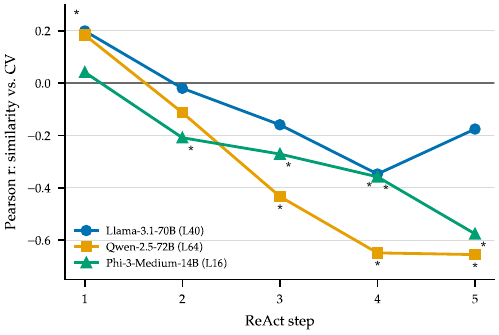}
\caption{Step progression at fixed readout layers: Llama L40, Qwen L64, and Phi-3 L16. Llama is strongest at step~4. Qwen's step-4 and step-5 point estimates are nearly tied ($-0.649$, $n=90$ vs.\ $-0.655$, $n=58$), so we do not assign it a unique temporal peak. Phi-3 is strongest at step~5. Eligible question sets shrink at later steps, so cross-step differences are descriptive. Asterisks mark unadjusted $p<.05$.}
\label{fig:step_progression_multimodel}
\end{figure}

Figure~\ref{fig:step_progression_multimodel} shows the temporal profile at one reported readout layer per model.

\section{Hard vs.\ easy breakdown}
\label{app:hardeasy}

\begin{figure}[H]
\centering
\includegraphics[width=\columnwidth]{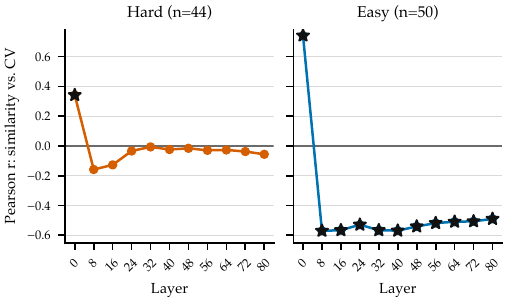}
\caption{Step-4 layer-wise correlation between activation similarity and behavioral CV, split by eligible group: hard $n=44$ and easy $n=50$. Six hard questions lack at least two valid step-4 states and are excluded. The plot shows where the association disappears but does not identify the cause. Stars denote unadjusted two-sided Pearson $p<.05$.}
\label{fig:hard_vs_easy}
\end{figure}

Figure~\ref{fig:hard_vs_easy} shows the step-4 layer-wise correlation split by question difficulty. On easy questions the similarity--CV relationship is strong and significant at every non-embedding layer ($r \approx -0.5$ to $-0.57$); on hard questions it is near zero ($|r|<0.16$ at all layers). This pattern is compatible with the correctness-agnostic account in Section~\ref{sec:boundaries}, but it does not identify the cause. The positive layer-0 points may reflect the lexical effects discussed in Appendix~\ref{app:crossmodel}.

\section{Baseline predictors of behavioral CV}
\label{app:baselines}

\begin{table}[H]
\centering
\small
\begin{tabular}{lrr}
\toprule
Predictor & $r$ & $p$ \\
\midrule
\textbf{Hidden-state sim.\ (S4, L40)} & $\mathbf{-0.35}$ & $\mathbf{.0006}$ \\
\midrule
TF-IDF obs.\ similarity$^\ddagger$ & $-0.63$ & $< .0001$ \\
Jaccard doc.\ similarity$^\ddagger$ & $-0.40$ & $< .0001$ \\
Search query overlap$^\ddagger$ & $-0.24$ & $.021$ \\
\midrule
Accuracy (correct rate)$^\dagger$ & $-0.31$ & $.002$ \\
Mean total thought length$^\dagger$ & $0.26$ & $.011$ \\
Mean search actions$^\dagger$ & $0.23$ & $.019$ \\
Mean step count$^\dagger$ & $0.21$ & $.037$ \\
\midrule
Question word count & $0.10$ & $.31$ \\
Question char count & $0.12$ & $.25$ \\
N context documents & $0.04$ & $.66$ \\
Mean thought length (S3) & $0.12$ & $.24$ \\
\bottomrule
\end{tabular}
\caption{Predictors of behavioral CV. $\dagger$Downstream behavioral consequences (post-completion), not pre-completion confounds. $\ddagger$Observation overlap measures computed across runs at steps 1--3 ($n = 94$).}
\label{tab:baselines_full}
\end{table}

\section{Prototype similarity signal for hard questions}
\label{app:prototype}

The primary measure is cross-run convergence within a question. We also consider a between-question measure: cosine similarity of each hard question's mean step-4, layer-32 state to the hard-question centroid. Higher prototype similarity is associated with higher CV ($r=0.43$, permutation $p=0.004$, bootstrap 95\% CI $[0.24,0.63]$, $d=0.64$, $n=44$).\footnote{Six of 50 hard questions are excluded because no run produced a step-4 hidden state. Excluded and included questions do not differ in accuracy ($p=.97$), but they do differ in CV ($p<.001$) because short trajectories mechanically have low step-count variance.} Equivalently, questions farther from the prototype tend to be more consistent. Variance of pairwise similarity ($r=-0.22$, $p=0.14$), trajectory slope ($r\approx0$), and within-run temporal stability ($r=0.28$, $p=0.06$) are not significant.

\begin{figure}[H]
\centering
\includegraphics[width=0.7\columnwidth]{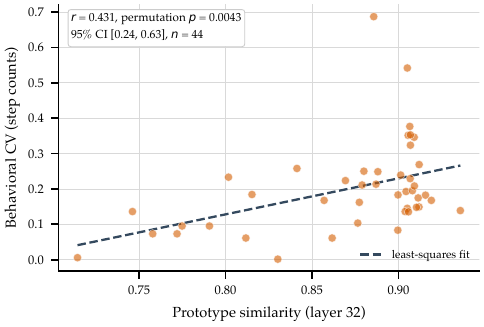}
\caption{Hard-question prototype similarity signal: cosine similarity to the hard-question centroid vs.\ behavioral CV (step~4, layer~32). Higher similarity is associated with higher CV ($r = 0.43$, permutation $p = 0.004$; $n=44$). The dashed line is the ordinary least-squares fit of the plotted points. Questions with CV exactly zero receive a small deterministic vertical jitter so that overlapping points remain countable.}
\label{fig:prototype}
\end{figure}

\section{Exploratory linear-direction analysis}
\label{app:geometry}

We explore whether the measured variation is approximately linear \citep{park2023linear}. At the primary Llama layer-40 readout, we define a direction from the difference between the high-accuracy group and the low-accuracy, higher-CV group. This direction nearly coincides with the first principal component of the same question states ($\cos=-0.98$; PC1 explains 53\% of variance). Its in-sample projection correlates with behavioral CV ($r=-0.32$, $p=0.001$; easy $r=-0.41$; hard $r=-0.34$). However, the direction is defined and evaluated on the same questions and strongly aligns with the easy--hard difference ($\cos=0.95$). It is therefore not evidence of a commitment-specific axis. The analogous Qwen projection at layer~64 has $r=-0.59$. A Procrustes-aligned cross-model comparison gives modest agreement ($\cos=0.19$; raw PCA-space $\cos=0.38$). These geometric results are exploratory and require held-out validation.

\begin{figure}[H]
\centering
\includegraphics[width=0.7\columnwidth]{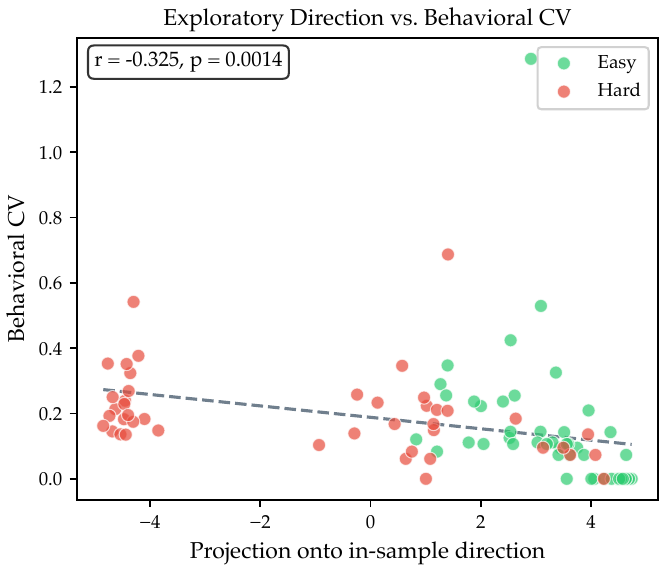}
\caption{Exploratory in-sample projection vs.\ behavioral CV at Llama layer~40, step~4. Easy questions are green and hard questions are coral. The direction is fit and evaluated on the same questions and strongly aligns with difficulty, so this figure does not establish a commitment-specific axis.}
\label{fig:commit_proj}
\end{figure}

\section{Consistency prediction: full results}
\label{app:runtime}

Table~\ref{tab:runtime_full} reports the full evaluation across features, models, and labeling schemes.

\begin{table}[H]
\centering
\small
\begin{tabular}{lcccc}
\toprule
 & \multicolumn{2}{c}{Llama 3.1 70B} & \multicolumn{2}{c}{Qwen 2.5 72B} \\
\cmidrule(lr){2-3} \cmidrule(lr){4-5}
Feature & Quintile & Median & Quintile & Median \\
\midrule
A: Sim (L40) & .93 & .55 & .87 & .86 \\
B: Sim (avg) & .95 & .58 & .90 & .88 \\
C: Layer profile & \textbf{.97} & .73 & .89 & .88 \\
D: Trajectory & .74 & .55 & .77 & .79 \\
E: PCA (L40) & .94 & \textbf{.85} & \textbf{.93} & .67 \\
\midrule
Question length (pre) & .52 & .60 & .65 & .57 \\
Context docs (pre)$^\S$ & .08 & .06 & .07 & .06 \\
Thought length (post)$^\S$ & .10 & .04 & .00 & .00 \\
\bottomrule
\end{tabular}
\caption{Consistency-prediction AUROC (LOOCV point estimates). Features A--E use step-4 hidden states; the layer-profile Llama quintile result has 95\% CI $[0.90,1.00]$. Surface rows show raw AUROC without post hoc inversion. Thought length is available only after completion. Context-document count is available earlier and has strong inverse association, but its direction was not selected inside each training fold; we therefore do not use it to claim superiority for hidden-state features.}
\label{tab:runtime_full}
\end{table}

\begin{figure}[H]
\centering
\includegraphics[width=0.98\columnwidth]{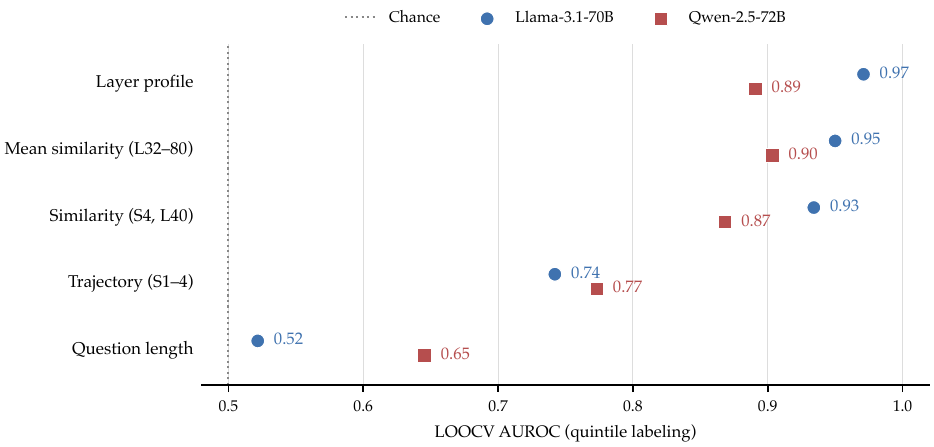}
\caption{Runtime-monitor AUROC by feature under quintile labeling. Hidden-state features perform strongly on both models; the layer-profile feature reaches 0.97 on Llama. Question length is shown as a simple pre-completion baseline. The PCA feature is omitted here for legibility, so Qwen's best quintile score ($0.93$) appears only in Table~\ref{tab:runtime_full}, which also discusses the unresolved directionality of context-document count.}
\label{fig:runtime_roc}
\end{figure}

\begin{figure}[H]
\centering
\includegraphics[width=0.7\columnwidth]{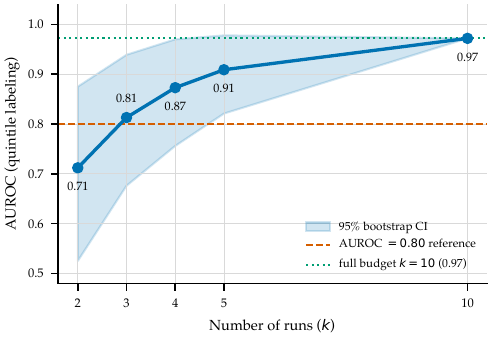}
\caption{Consistency-prediction AUROC vs.\ number of runs. At $k=3$, mean AUROC is $0.81$ with subsampling SD $0.07$. Shading shows the stored 95\% bootstrap confidence interval, verified to bracket the mean at every $k$. The dashed line marks AUROC $0.80$ as an interpretive high-discrimination reference; the dotted line is the full-budget $k=10$ estimate.}
\label{fig:auroc_k}
\end{figure}

\textbf{Fewer runs.} Performance decreases as the run budget shrinks. With $k=3$ runs, mean AUROC is $0.81$ (subsampling SD $0.07$; 95\% bootstrap CI $[0.68,0.94]$). It rises to $0.87$ at $k=4$, $0.91$ at $k=5$, and $0.97$ at $k=10$ (Figure~\ref{fig:auroc_k}). At $k=3$, precision is $0.81$ at recall $0.90$. A new task would still require calibration of the readout step and sampled layers.

\section{StrategyQA cross-benchmark results}
\label{app:strategyqa}

Table~\ref{tab:strategyqa_layers} reports layer-wise correlations at step~3 (the peak step) for Llama-3.1-70B on StrategyQA ($n{=}50$). All layers 8--80 are significant at $p < 10^{-8}$; partial correlations controlling for accuracy are virtually unchanged given the high accuracy (93.2\%).

\begin{table}[H]
\centering
\footnotesize
\setlength{\tabcolsep}{4pt}
\begin{tabular}{lcccc}
\toprule
Layer & Raw $r$ & Raw $p$ & Partial $r$ & Partial $p$ \\
\midrule
0  & -- & -- & -- & -- \\
8  & $-0.73$ & $2.5 \times 10^{-9}$ & $-0.72$ & $2.9 \times 10^{-9}$ \\
16 & $-0.78$ & $3.8 \times 10^{-11}$ & $-0.77$ & $4.3 \times 10^{-11}$ \\
24 & $-0.76$ & $2.3 \times 10^{-10}$ & $-0.75$ & $3.7 \times 10^{-10}$ \\
32 & $-0.78$ & $2.6 \times 10^{-11}$ & $-0.78$ & $4.0 \times 10^{-11}$ \\
40 & $-0.81$ & $8.7 \times 10^{-13}$ & $-0.81$ & $8.4 \times 10^{-13}$ \\
48 & $-0.82$ & $2.0 \times 10^{-13}$ & $-0.82$ & $1.9 \times 10^{-13}$ \\
56 & $-0.83$ & $8.8 \times 10^{-14}$ & $-0.83$ & $9.7 \times 10^{-14}$ \\
64 & $-0.83$ & $7.7 \times 10^{-14}$ & $-0.83$ & $9.3 \times 10^{-14}$ \\
\textbf{72} & $\mathbf{-0.83}$ & $\mathbf{7.4 \times 10^{-14}}$ & $\mathbf{-0.83}$ & $\mathbf{9.1 \times 10^{-14}}$ \\
80 & $-0.82$ & $4.8 \times 10^{-13}$ & $-0.82$ & $4.7 \times 10^{-13}$ \\
\bottomrule
\end{tabular}
\caption{StrategyQA: layer-wise correlations at step~3 (Llama-3.1-70B, $n{=}50$). The signal spans all non-embedding layers and peaks at layer~72, plateauing across layers 56--72. Partial correlations control for accuracy.}
\label{tab:strategyqa_layers}
\end{table}

\begin{table}[H]
\centering
\footnotesize
\setlength{\tabcolsep}{3pt}
\begin{tabular}{lcc}
\toprule
 & HotpotQA & StrategyQA \\
\midrule
Task type & Multi-hop retrieval & Implicit reasoning \\
Answer type & Span extraction & Yes/no \\
$n$ (eligible at readout) & 94 & 50 \\
Runs per question & 10 & 10 \\
Mean chain length & ${\sim}12$ steps & 5.1 steps \\
Accuracy & 70\% & 93.2\% \\
\midrule
Reported step & 4 & 3 \\
Readout layer & 40 & 72 \\
$r$ at readout & $-0.35$ & $-0.83$ \\
95\% CI & $[-0.51, -0.16]$ & $[-0.90, -0.72]$ \\
Peak $p$ & $< 0.001$ & $< 10^{-13}$ \\
Partial $r$ & $-0.45$ & $-0.83$ \\
\bottomrule
\end{tabular}
\caption{Cross-benchmark comparison at the reported Llama-3.1-70B readouts. StrategyQA has a stronger association at step~3 than the primary HotpotQA layer-40 association at step~4.}
\label{tab:crossbenchmark}
\end{table}

\begin{figure}[H]
\centering
\includegraphics[width=0.80\columnwidth]{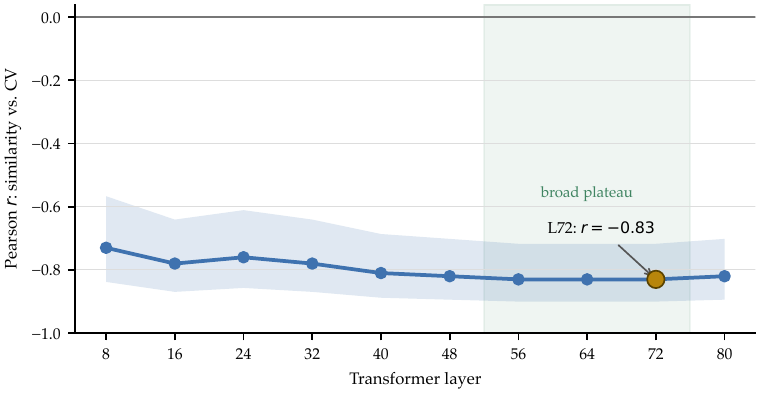}
\caption{Visualization of the StrategyQA step-3 values reported in Table~\ref{tab:strategyqa_layers} ($n=50$); shading gives Fisher-$z$ intervals. The per-question StrategyQA hidden states were not retained, so this panel visualizes the tabulated correlations and is not an independent reanalysis.}
\label{fig:strategyqa_heatmap}
\end{figure}

\section{Summary of all results}
\label{app:summary}

\begin{table*}[ht]
\centering
\scriptsize
\setlength{\tabcolsep}{3pt}
\resizebox{\textwidth}{!}{%
\begin{tabular}{llccccccc}
\toprule
\textbf{Model} & \textbf{Benchmark} & $n$ & \textbf{Step} & \textbf{Readout layer (\% depth)} & \textbf{$r$} & \textbf{Partial $r$} & \textbf{Best AUROC} & \textbf{CV red.} \\
\midrule
Llama-3.1-70B & HotpotQA   & 100 & 4 & 40 (50\%) & $-0.35$ & $-0.45$ & 0.97 & 28\% \\
Qwen-2.5-72B  & HotpotQA   & 100 & 4 & 64 (80\%) & $-0.65$ & --     & 0.93 & -- \\
Phi-3-Medium-14B & HotpotQA & 100 & 5 & 16 (40\%) & $-0.58$ & --     & --  & -- \\
Llama-3.1-70B & StrategyQA  &  50 & 3 & 72 (90\%) & $-0.83$ & $-0.83$ & --  & -- \\
\bottomrule
\end{tabular}
}
\caption{Summary at the reported readouts. Here $n$ is the number of sampled questions; 94 Llama-HotpotQA questions are eligible for the primary step-4 analysis, and 91 Phi-3 questions are eligible at step~4. The Phi-3 row reports its strongest temporal result at step~5; the comparable step-4/L16 value is $r=-0.36$ (Table~\ref{tab:crossmodel_full}). Best AUROC is the best quintile-labeled feature for each model (full results in Table~\ref{tab:runtime_full}); CV reduction is Commitment vs.\ Filler.}
\label{tab:summary}
\end{table*}

\section{Intervention details}
\label{app:intervention_decomposition}

\textbf{Interpreting the three comparisons.} The filler condition trends toward \emph{higher} variance than control (CV $= 0.132$ vs.\ $0.112$, $+18\%$, $d = 0.18$, $p = .071$), suggesting any prompt insertion at step~3 disrupts the natural trajectory. The commitment prompt overcomes this disruption \emph{and} tightens beyond baseline (15\% net CV reduction, $d = 0.20$, $p = .044$, Holm $p = .088$). Reading the three together: (i) Filler vs.\ Commitment (28\%, $d = 0.33$, Holm $p = .003$) isolates framing from token count; (ii) Control vs.\ Commitment (15\%, $d = 0.20$, Holm $p = .088$) is the net behavioral effect over the standard agent, which does not survive correction; (iii) Control vs.\ Filler ($+18\%$, n.s.) is the disruption cost.

\textbf{Mediation analysis.} For the Filler-to-Commitment contrast, a bootstrap mediation analysis \citep{preacher2008asymptotic} (5{,}000 iterations, $n=89$ matched questions) gives an indirect path through activation similarity of $ab=-0.062$ (95\% CI $[-0.094,-0.038]$, $p<.001$). The direct path has the opposite sign and is not significant ($c'=+0.021$). This is inconsistent mediation, or suppression, rather than full mediation: the total effect is $c=ab+c'=-0.041$. We therefore report a statistical decomposition, not a causal mechanism.

\begin{figure}[H]
\centering
\includegraphics[width=0.92\columnwidth]{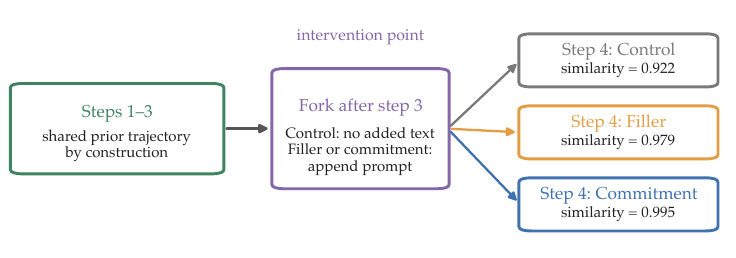}
\caption{Intervention timing and the first post-intervention comparison. All conditions share the prefix through step~3. Control then receives no added text; Filler and Commitment receive their respective prompts. Layer-40 similarity is measured at step~4 ($n=89$): Commitment 0.995, Filler 0.979, Control 0.922.}
\label{fig:intervention_step_progression}
\end{figure}

\textbf{Time-locking.} Figure~\ref{fig:intervention_step_progression} makes the experimental timing explicit. The three conditions share the same prefix through step~3. The prefix is then forked: Control receives no added text, while Filler and Commitment receive their respective prompts. Activation similarity is compared at step~4, so the first reported difference cannot reflect different question subsets.

\textbf{Stratified analysis.} The effect is largest for questions in the top tertile of Control-condition CV ($n=32$): Filler vs.\ Commitment gives $\Delta\text{CV}=0.069$ ($d=0.44$, $p=.018$). Questions with lower Control-condition CV show a smaller, non-significant trend ($d=0.27$, $p=.102$).

\textbf{Question-level decomposition.} Classifying each question under both conditions into consistent-correct (accuracy $\geq 0.8$), consistent-wrong (accuracy $\leq 0.2$, majority frequency $\geq 0.6$), and inconsistent yields a stable 3$\times$3 transition matrix over all 100 questions. Under Control the split is 89 consistent-correct, 4 consistent-wrong, and 7 inconsistent. Moving to Commitment, 88 of the 89 consistent-correct questions stay so, all 4 consistent-wrong stay so, and none move from consistent-correct to consistent-wrong. This descriptive sample contains no observed consistent-correct to consistent-wrong transition; it is not a guarantee that commitment cannot degrade a correct trajectory.

\begin{figure}[H]
\centering
\includegraphics[width=0.7\columnwidth]{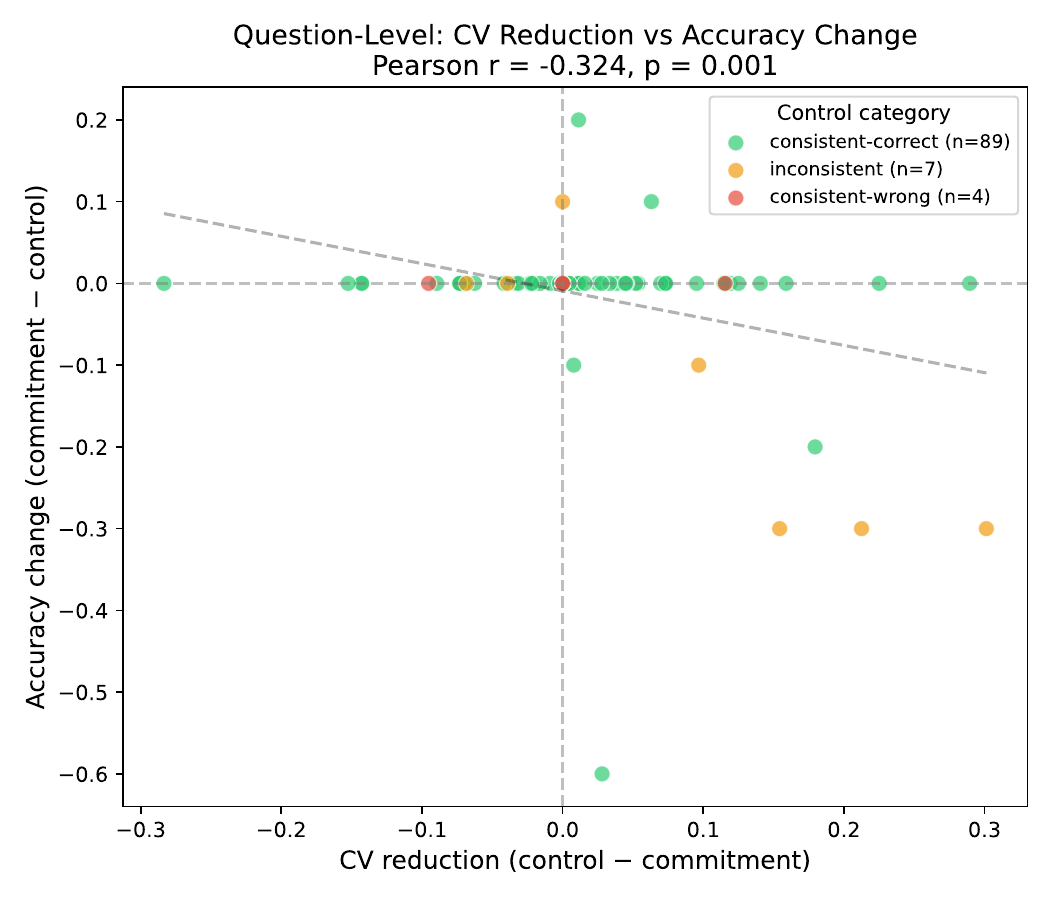}
\caption{Question-level CV reduction ($x$) vs.\ accuracy change ($y$) under commitment ($n=100$). The displayed points yield $r=-0.32$ ($p=.001$); the dashed line is their linear fit. Transparency and deterministic vertical jitter reveal the many questions with no accuracy change.}
\label{fig:cv_acc_scatter}
\end{figure}

Figure~\ref{fig:cv_acc_scatter} relates CV reduction to accuracy change at the question level. The negative correlation ($r = -0.32$, $p = .001$) shows commitment amplifies the existing trajectory: questions where CV dropped most tended to show slight accuracy decreases, consistent with the correctness-agnostic commitment signal (\S\ref{sec:boundaries}).

\section{Intervention prompt text}
\label{app:prompts}

\textbf{Commitment prompt} (appended at step~3): ``Based on the evidence you have gathered so far, commit to a specific reasoning strategy for solving this question. State your committed strategy clearly in your next Thought, then follow through with it. Do not change strategies or start over; build on what you have learned.''

\textbf{Filler prompt} (appended at step~3, matched token length): ``Please continue with the task as you normally would. Take the time you need to work through the problem. Consider the information you have gathered and proceed with your next step in whatever way seems most appropriate to you. There is no particular urgency; work at your own pace and follow your reasoning.''

\end{document}